# Guided inverse design of surface plasmon polaritons based nanophotonic films via deep learning

**Yingshi Chen[1], Jinfeng Zhu[1] and Qing Huo Liu[2]**

[1] Institute of Electromagnetics and Acoustics, and Department of Electronic Science, Xiamen University, Xiamen 361005, China
E-mail: nanoantenna@hotmail.com (J Zhu)
[2] Department of Electrical and Computer Engineering, Duke University, Durham, North Carolina 27708, USA

## Abstract

In this paper, We present an improved deep convolutional neural network, Guided-Resnet, to design low-cost surface plasmon polaritons (SPP) based nanophotonic films. The input of Guided-Resnet is the two dimensional spectrum map that has the optical response required by the user. Two dimensional map has much more information and features than one dimensional curve. The deep CNN has powerful ability to extract features and learn models from these maps. We use the guided sample replacement algorithm in the training process, which would gradually learn from samples with lower cost. The experiment shows that the average relative accuracy of spectrum is higher than 90%. More importantly, the cost of predicted structure is much lower than standard deep CNN. Our work shows that deep learning would not only design multi-layer films with high accuracy but also learn how to reduce the cost. That is, our model learned to replace precious metals with ordinary metals.

## 1. Introduction

Surface plasmon polaritons (SPP) are two-dimensional electromagnetic waves confined at the metal–dielectric interfaces [1]. They can be used to break the diffraction limit, which is very important for many areas such as information storage ,biosensing, and integrated photonic circuits [2]. But the inverse design of SPP structures is an ill-posed inverse problem[3]. There are many-to-one correspondence between the structures and user needs. Users not only want to get a solution, but also need the solution to be optimal on some criterion or some cost function. This is the key problem of inverse design process.

We mainly investigate low-cost multilayer metal/dielectric nanocomposite films that support surface plasmon polaritons. Usually users want SPP at specified incident angle and wavelength, which could be described by two dimensional heat maps of reflectance. The most used metal types are gold (Au), silver (Ag), copper (Cu) and aluminium (Al). The cost of Au is much expensive than Ag, Cu and Al. So not only the films' spectrum should meet the users' requirements, but the lower the cost, the better. That is,user want to design films with as less au as possible. Recently, machine learning methods showes great power in design nano-photonics[4-9]. But all these models do not take cost factor into account. The predicted design result may have much higher cost than the optimal solution. No other teams find ways to tackle the cost problem by deep learning. In this paper, we present novel guiding mechanism for this problem, which will guide convolutional neural network from training samples with lower and lower cost. And it not only predicts a solution with high accuracy, but also find much lower cost solution.To our knowledge, It's a first step in the low cost inverse design problem by deep learning.

The reverse design of nanophotonic devices has been widely studied in the last year [13-24]. [13] trained a bidirectional network to design the nanoparticle geometry by the optical response spectrum. [15] studied three types of plasmonic NPs including nanosphere, nanorod, and dimer. They establish mappings between the far-field spectra/near-field distribution and dimensional parameters. [16] approximate light scattering by multilayer nanoparticles and solve nanophotonic inverse design problems by using back propagation. [17] used machine learning for the light-matter interaction problem, which governs those fields involving materials discovery, optical characterizations, and photonics technologies. [18] used the dimensionality reduction technique to significantly reduce the dimensionality of a generic EM wave-matter interaction problem without imposing significant error. [19] find a deep learning approach to improve detection characteristics of surface plasmon microscopy (SPM) of light scattering. [20] studied the problem of 2-bit programmable coding metasurface by machine learning techniques to realize a ground breaking imaging hardware platform. [21] reported that design optimizations can be significantly sped-up when paired with deep neural networks. [22] propose a novel learning strategy based on neuroevolution to design and train the ONNs. [23] find that artificial neural networks provides a powerful and efficient tool to construct accurate correlation between plasmonic geometric parameters and resonance spectra.

Compared to recent works [13-24] of other teams, we use two dimensional heat maps instead of one dimensional spectra curves, which are used by nearly all other teams. To our knowledge, it's the first attempt to use 2D map as training samples. 2D map have much more details. More importantly, we can use deep convolutional neural networks (CNN). Although all teams adopt the framework of deep learning, most other teams use classical deep multi-layer perceptions (MLP). While we use deep CNN to train these 2D maps. Deep CNN has much stronger recognition and discrimination ability than classical MLP. Since CNNs has convolution and pooling layers to exploit spatial invariance. If we use MLP to train image, the images have to be flattened and lost spatial information. Also, CNNs share weights in the convolutional process to cut down on the number of parameters, which makes them extremely efficient in image processing, compared to MLPs.

We also have made breakthroughs in the following two novel points. 1) We use novel guided algorithm in training processes, which would work for any user-need cost functions. So this method would meet users' guideline. The author of [14] raised an important question: "The lack of design guidelines for data-driven methods to deal with." Our novel guided algorithm is the first attempt to answer this question and get good results. 2) We use hybrid network to solve both classification and regression problem. Similar scenes are often encountered in the inverse design of nanophotonics structures. For example, [15] employed another network to classify the spectra of different nanostructures. Single hybrid network would reduce the training time and the size of model greatly compare to two independent networks. Our approach showed the potential of hybrid network, which could apply to other similar problems.

## 2. Methodology

To verify the guided deep learning method, we use a specific structure showned in Figure 1.It's a multilayer metal/dielectric nanocomposite films, and the excitation light is introduced through a hemispheric prism with a incident angle $\theta$. The purpose of the inverse design is to find the thickness and metal type of eah layer to generate user needed spectrum.

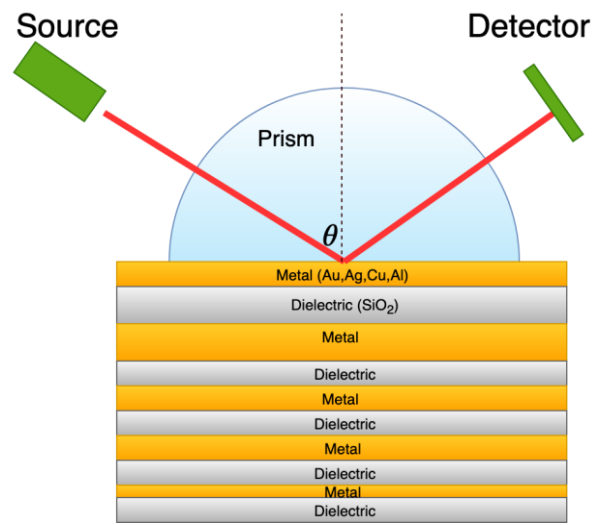

Figure 1

## 2.1. Two dimensional heat maps

We use deep CNN to learn 2D heat maps of SPP films. This is the one of main difference between us and [4-9, 13-24]. All backbones of neural netwoks in [4-9, 13-24] are classical multilayer perceptrons (MLP) and the inputs are 1D spectrums. Our deep CNN uses many small convolution kernels to extract the subtle and distinguishing features. The stacked pooling layers and more complex modules would get more hierarchical features at the different scales. So CNN has much more powerful learning ability than simple MLP. Figure 2 showes some 2D heat maps of SPP multilayer films, which generated by the characteristic matrix method [10]. We can see the dips, lengths and sharp edge of SPP curves in the heat map.There are also regions with sudden change and regions with slow dispersion. All these subtle and rich detail information could be analyzed by deep CNN.

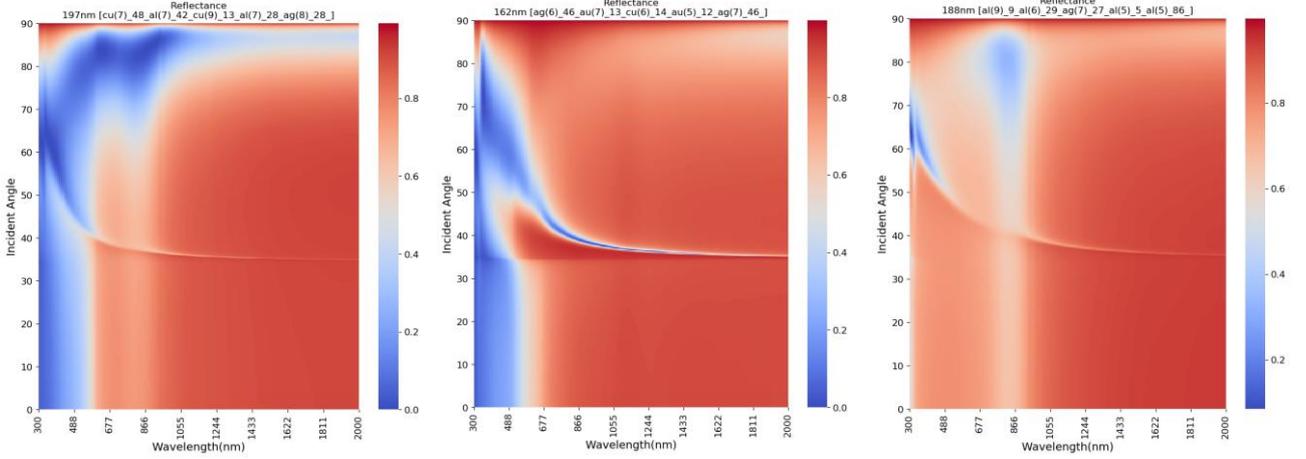

Figure 2. Some 2D heat maps of SPP multilayer films. We can see the dips, lengths and ranges of SPP curves. The X axis is the wavelengths from 300nm to 2000nm and Y axis is the incident angle $\delta$ from 0° to 90°

The generation process of heat maps is as following.Given the material properties $\eta$ and incident angle $\delta$, we can get the characteristic matrix for each intermediate layer:

$$\bar{M} = \begin{bmatrix} \cos\delta & \left(j/\eta\right)\sin\delta \\ j\eta\sin\delta & \cos\delta \end{bmatrix} \qquad (1)$$

All intermediate matrices are then multiplied to get the total matrix:

$$\bar{M}_{total} = \bar{M}_2\bar{M}_3\cdots\bar{M}_{k-1} \qquad (2)$$

Let $\bar{M}_{total} = \begin{bmatrix} m_{11} & m_{12} \\ m_{21} & m_{22} \end{bmatrix}$, then the reflection coefficientof the entire multiple-layer structure is

$$\rho = \frac{\eta_1(m_{11}+m_{12}\eta_k)-(m_{21}+m_{22}\eta_k)}{\eta_1(m_{11}+m_{12}\eta_k)+(m_{21}+m_{22}\eta_k)} \qquad (3)$$

For each incident angle $\delta$ in the range[0 ° ~ 90 °], we get 1D reflectance curve $\tilde{A}$ at different wavelengths from 300nm to 2000nm. Then we merge these curves along Y axis to generate 2D heat maps.

## 2.2. Guided training of CNN with hybrid loss

In the case of on-demand design of low-cost multilayer films, users hope to get a low cost structure with correct optical response. We solve this proble in the framework of deep learning. The input $O_i$ is the spectrum map that has the optical response required by the user. The output $Y_i$ is the structure paramters vector, which represents the thickness and type of each layer. We trained a novel Guided-Resnet network for this problem. As shown in figures 3, the backbone of Guided-Resnet is classical residual network Resnet-18[11,12], which has powerful representational ability to extract features from spectrum map. At the input layer, each map $O_i$ is resized to 224x224 with 3 color channels. Then passed to a layer with 64 convolution kernels(7x7). The next 4 layers perform 3x3 convolution with [64, 128, 256, 512] kernels respectively. The final feature vector at the full connection layer has 512 values. At the outpt layer, we get the structure paramters $Y_i$. A hybrid loss function would get its' loss value and the gradient for back propagation. The dashed blue curve corresponds to the guided learing process, which will lower down the cost. We give detailed description below.

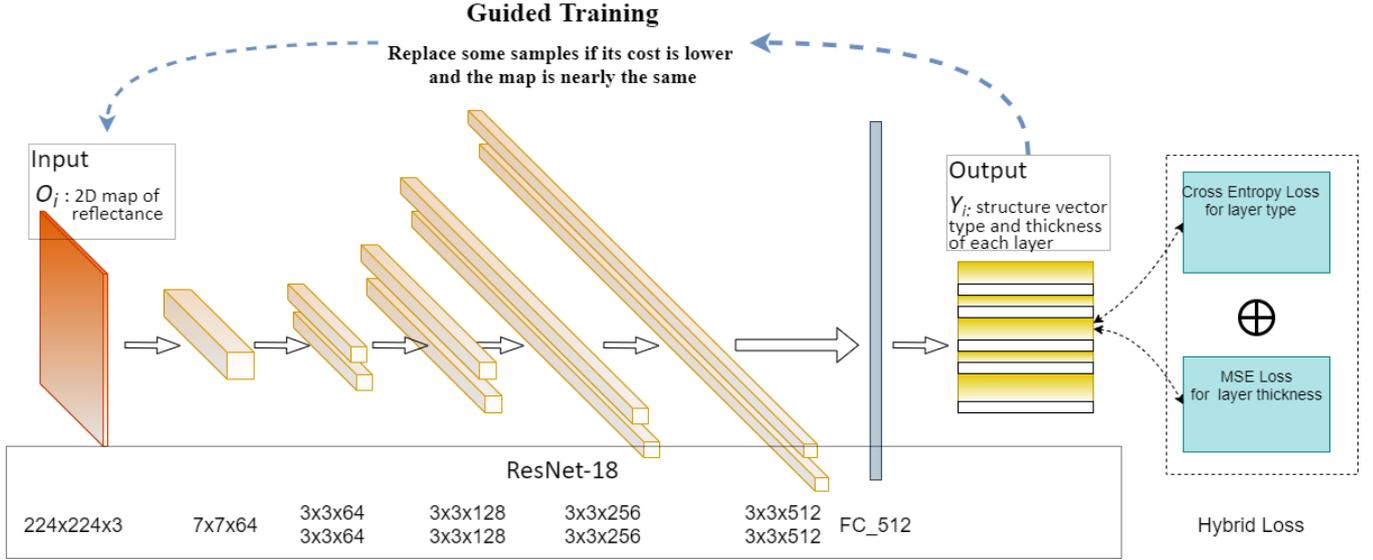

Figure 3 the structure of Guided-Resnet.

**1) Low cost sample replacement algorithm**

Only with Resnet, the learned model does not take price into account. 在这种情况下，The design results may be too expensive to be accepted by customers. In order to solve this problem, we enhance Resnet-18 with guiding mechanism to find low cost solutions.We present an low cost sample replacement algorithm to find optimal solutions. For each N-layer sample i in training set, formula (1) define its structure vector $Y_i$ , which includes the thickness $h^l$ and material type $m^l$ of each layer. The dielectric $m^l$ at metal layer is always one of Au,Ag,Cu and Al.

$$Y_i = \{h^l, m^l\} \;\; l = 1,2,\cdots,N \tag{1}$$

Based on the structure vector $Y_i$, we define the cost $c_i$ as formula (2).The $cost(m^l)$ is user-specified function on the material type. For example, the price of metal.

$$c_i = \sum_l h^l \times cost(m^l) \;\; l = 1,2,\cdots,N \tag{2}$$

From the structure vector $Y_i$, we generate the 2D map $O_i$ by the the characteristic matrix method.

In each step of training process, the network would generate new structure vector $\tilde{Y}_i$ ,which is the result of output layer at the current step. Generally, the corresponding structure of $Y_i$ and $\tilde{Y}_i$ is different. The corresponding cost is also different.We use the following algorithm to pick low cost sample. $O_i$ $Y_i$

Algorithm 1 Low cost sample replacement algorithm

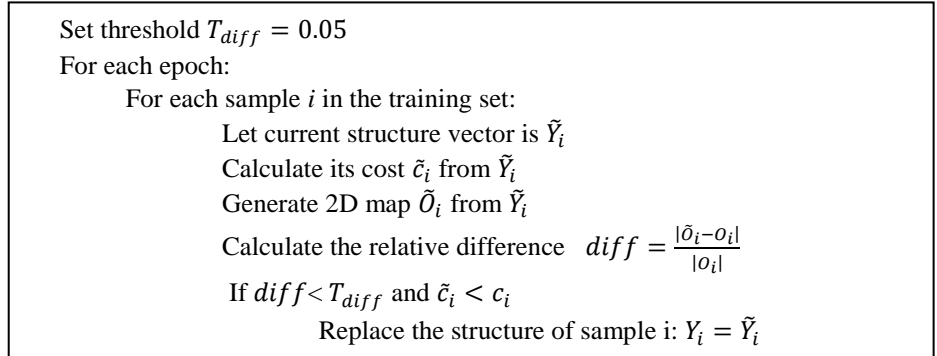

In the training process, we would generate new 2D map $\tilde{O}_i$ and its cost $\tilde{c}_i$ from $\tilde{Y}_i$ . If $\tilde{O}_i$ nearly the same as $O_i$, and $\tilde{c}_i$ is lower than current $c_i$, we repalce its original structure $Y_i$ with $\tilde{Y}_i$. At each training epoch, some samples would be replaced by low cost alternatives. As the training goes on, the average cost of training set will gradually decline, while spectrum remains essentially same. With the gradually declining cost of training samples, the cost of prediction samples would lower and lower. Then we can get much better solution, which spectrum has same accuracy and its cost much lower than classical network.

**2) Hybrid loss**

For each sample i , the target $Y_i$ represents thickness and material type of each layer, to compare the loss between $Y_i$ and $\tilde{Y}_i$(output of network) , we use a hybrid loss function as following:

$$Loss = Loss(thickness) + Loss(material\ type) \quad (3)$$

The loss function of thickness is defined as follows:

$$Loss(thickness\ ) = \frac{\sqrt{\sum_{i=1}^{n}(t_i - \tilde{t}_i)^2}}{\sqrt{\sum_{i=1}^{n} t_i{}^2}} \quad (4)$$

where $t_i, \tilde{t}_i$ is the target thickness and predicted thickness of each layer

The loss function of material type is defined as follows:

$$Loss(material\ type) = 1 - (softmax(\tilde{m}_i)\ equal\ to\ m_i) \quad (5)$$

where $m_i$ is the target material type of each layer. We use softmax function[25] to predict the layer type from the output $\tilde{m}_i$ of network.

We can see that loss of thickness is regression metric and loss of material type is a classification metric. So in some sense, the Guided-Resnet is a hybrid CNN. One regression CNN to predict the thickness and other classification CNN to predict the material type. And two CNNs share the same backbone(Resnet-18) to extract features.

For more details, the latest package can be accessed by the website https://github.com/closest-git/MetaLab. We hope this tool kit would facilitate the optical engineering and application of nanophotonics devices.

## 3. Results and discussion

To train the Guided Resnet, we first generate 5000 samples as the training set. Each sample has 10 metal/dielectric layers. The types of metal layers are random selected from gold (Au), silver (Ag), copper (Cu) and aluminium (Al). The material of dielectric layer is all SiO2. To get the two-dimensional map, we use 170 evenly distributed wavelengths from 300nm to 2000nm and 900 incident angle from 0˚to 90˚. So 153,000 curves for each map. We also generate 1000 additional samples as the testing set, which are different from the training set. We train the network by Adam method [26] with momentum 0.9. We use a batch size of 64 and learning rate is 0.01.

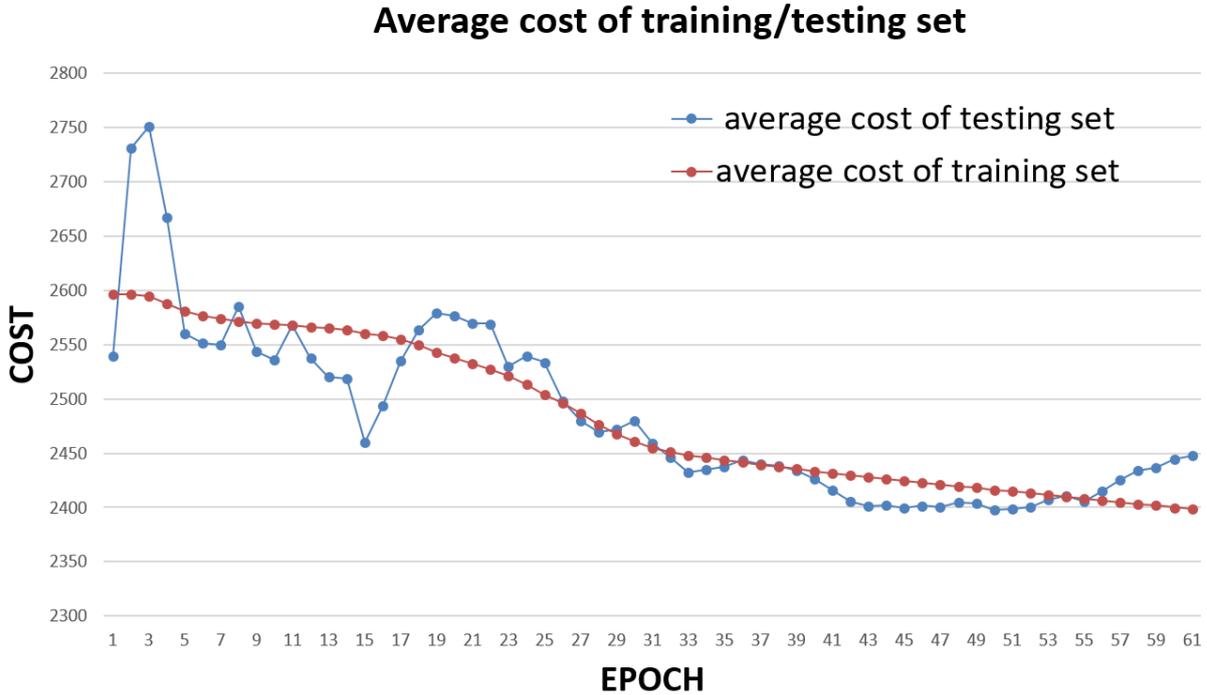

Figure 4 Average cost of training/testing set in the training process.
Each red point is the average cost of 5000 training samples.  Each blue point is the average cost of 1000 testing samples.

Figure 4 shows the change of average cost in the training process. Each red point is the average cost of 5000 training samples. Each blue point is the average cost of 1000 testing samples. The red curve is the average cost of training set. As the training going on, the cost gradually dropped from 2600 to 2400. The average cost of the testing set (the blue curve) is also declining around the red line. This verified that the guiding mechanism would predict the samples with lower cost. There are always biases in the prediction. So there are some oscillations in the cost of testing sets, which verified by the oscillation of blue curve.

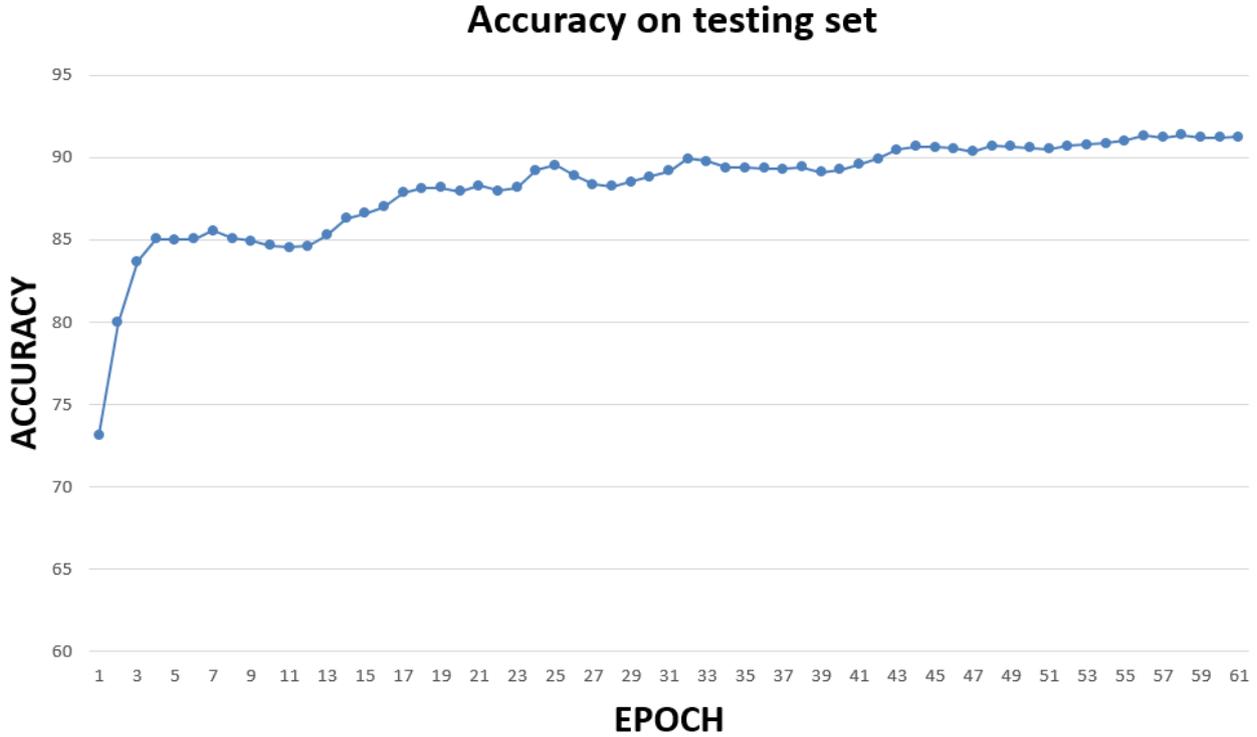

**Figure 5** Accuracy on testing set in the training process.
Each blue point is the average accuracy of 1000 testing samples.

Figure 5 shows the accuracy on the testing set. We use formula (6) to measure the accuracy:

$$\text{accuracy} = \frac{1}{N} \sqrt{\sum_{i=1}^{N} \frac{|\hat{S}-S|}{|S|}} \qquad (6)$$

where N is the size of testing set, $S$ is the 2D spectrum needed by use, and $\hat{S}$ is the predicted 2D spectrum. At each epoch in the training process, we use trained model to predict the spectrum $\hat{S}$, then compare to the used need spectrum $S$. After 45 epochs, the accuracy is always greater than 90%. Since the 1000 testing samples are generated by some random parameters. They are totally different from the training set. So this verified that our model would get accuracy higher than 90%.

Finally, we demonstrate two examples. Table 1 lists target and prediction results of the first sample,which has 10 layers. The "Target" column lists the structure parameters needed by user. The central column lists the result predicted by CNN without guiding mechanism. The right column lists the result predicted by CNN with guiding mechanism.The key difference is in layer 5. With guiding mechanism, the learned model replaces the expensive Au by cheaper Ag. The last row of table is the cost of each structure, which is calculated on the price of metals. So with guiding mechanism , the cost of predicted structure is only 11% of target device. Without guiding mechanism, the cost of predicted structure is 133% of target device. Figure 5 compares the prediction results of standard Resnet and our Guided Resnet. We can see both predictions have high accuracy. In these maps, most regions have the same spectral distribution. Only some difference around the curves of SPP.

Table 1 Comparison of structure and cost of sample 1

| Layer | Target | Prediction by Resnet | Prediction by Guided-Resnet |
|-------|--------|----------------------|------------------------------|
| 1 | Al - 5nm | Al - 7nm | Al - 5nm |
| 2 | SiO2 – 7nm | SiO2 – 9nm | SiO2 – 9nm |
| 3 | Ag - 6nm | Ag - 8nm | Ag - 5nm |
| 4 | SiO2 – 11nm | SiO2 – 20nm | SiO2 – 14nm |

| 5 | Au - 6nm | Au - 8nm | Ag - 5nm |
|---|---|---|---|
| 6 | $SiO2 - 8nm$ | $SiO2 - 14nm$ | $SiO2 - 18nm$ |
| 7 | Ag - 8nm | Ag - 8nm | Ag - 5nm |
| 8 | $SiO2 - 32nm$ | $SiO2 - 19nm$ | $SiO2 - 19nm$ |
| 9 | Al - 9nm | Al - 7nm | Al - 5nm |
| 10 | $SiO2 - 37nm$ | $SiO2 - 35nm$ | $SiO2 - 32nm$ |
| cost | 1996 | 2656 | 226 |

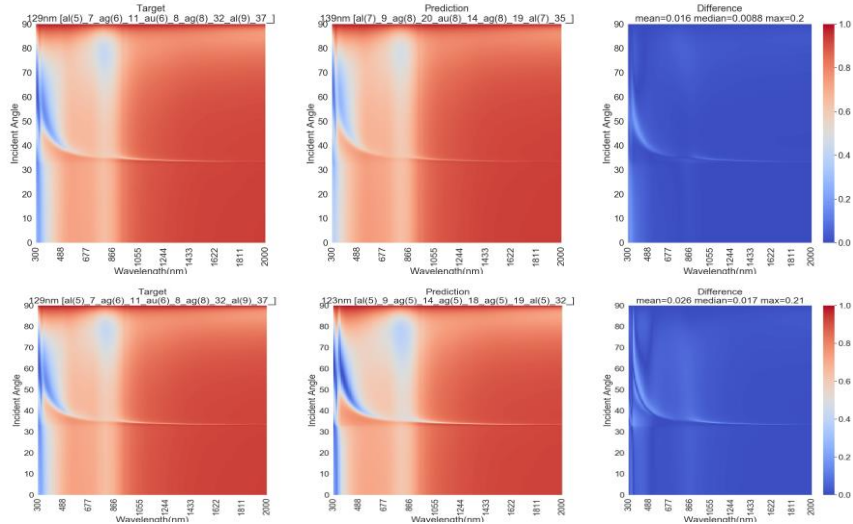

**Figure 5** Comparison of spectrums of sample 1.

The first row shows the results by Resnet; the average relative error is 1.6%. The second row shows the results by Guided-Resne; the average relative error is 2.6%. In both rows, left figure shows the target(user-need spectrums). The middle figure shows the predicted spectrums. The right figure shows the difference.

Table 2 lists the results of second sample. As we can see, the key difference is in layer 3. With guiding mechanism, the learned model replaces the expensive Au by cheaper Cu. So the cost is only 45% of target device. Without guiding mechanism, the cost of predicted structure is 105% of target device. Figure 6 compares the prediction results of standard Resnet and our Guided Resnet. We can see both predictions have high accuracy.

Table 2  Comparison of structure and cost of sample 2

| Layer | Target | Prediction by Resnet | Prediction by Guided-Resnet |
|---|---|---|---|
| 1 | Cu - 7nm | Au - 8nm | Cu - 5nm |
| 2 | $SiO2 - 9nm$ | $SiO2 - 7nm$ | $SiO2 - 14nm$ |
| 3 | Au - 8nm | Cu - 8nm | Cu - 5nm |
| 4 | $SiO2 - 26nm$ | $SiO2 - 32nm$ | $SiO2 - 18nm$ |
| 5 | Au - 7nm | Au - 8nm | Au - 5nm |
| 6 | $SiO2 - 14nm$ | $SiO2 - 11nm$ | $SiO2 - 19nm$ |
| 7 | Au - 7nm | Au - 7nm | Au - 5nm |
| 8 | $SiO2 - 17nm$ | $SiO2 - 15nm$ | $SiO2 - 17nm$ |
| 9 | Cu - 5nm | Cu - 6nm | Cu - 5nm |
| 10 | $SiO2 - 12nm$ | $SiO2 - 12nm$ | $SiO2 - 15nm$ |
| cost | 6601 | 6901 | 3001 |

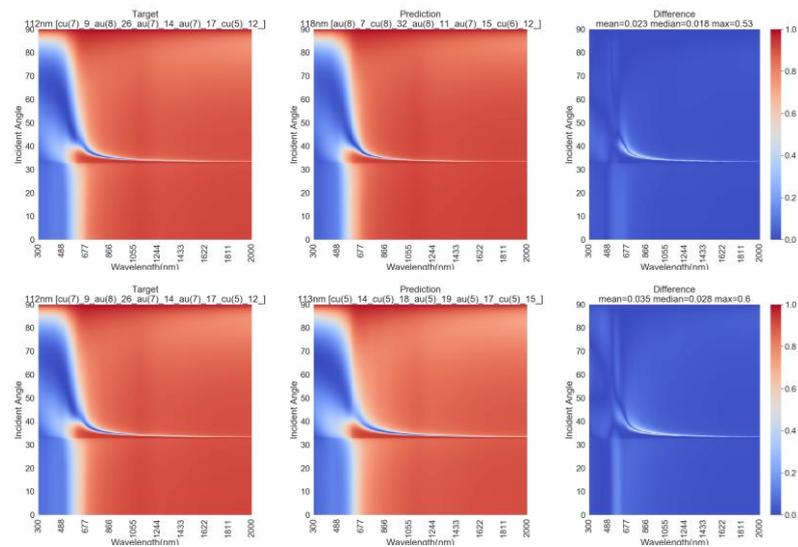

**Figure 6.** The first row shows the results by Resnet; the average relative error is 2.3%.The second row shows the results by Guided-Resnet; the average relative error is 3.5%. In both rows, left figure shows the target(user-need spectrums). The middle figure shows the predicted spectrums. The right figure shows the difference.

## 4 Conclusion and Prospect

We present a novel guiding mechanism for the inverse design of SPP devices. Comparing with user requirements, the average relative error of spectrum is less than 10%. More importantly, the cost of predicted structure is much lower than standard deep CNN.

The main difficulty in practical inverse design problems comes from the small sample problem. There are only several hundred or even fewer samples in many real applications. But deep learning is essentially a learning method based on huge samples. Its performance would decrease drastically with only a few samples. So we would use deep differentiable forest [27] for the small sample problem. Differentiable forest has the advantages of both trees and neural networks. We would use its' full differentiability to training and learning just like deep CNN. And its' tree model has more generality than the classical deep CNN. So it has the potential to solve inverse design problem with only 1000 samples or less.